\documentclass[conference]{IEEEtran}
\IEEEoverridecommandlockouts
\usepackage{cite}
\usepackage{amsmath,amssymb,amsfonts}
\usepackage{algorithmic}
\usepackage{graphicx}
\usepackage{textcomp}
\usepackage{xcolor}
\usepackage{booktabs}

\usepackage[T1]{fontenc} 

\usepackage{epstopdf}

\def\BibTeX{{\rm B\kern-.05em{\sc i\kern-.025em b}\kern-.08em
    T\kern-.1667em\lower.7ex\hbox{E}\kern-.125emX}}
\begin{document}

\title{Autoencoder based Randomized Learning of Feedforward Neural Networks for Regression\\
\thanks{Supported by Grant 2017/27/B/ST6/01804 from the National Science Centre, Poland (research) and and Grant 020/RID/2018/19 "Regional Initiative of Excellence" from the Minister of Science and Higher Education, Poland (conference).}
}

\author{\IEEEauthorblockN{Grzegorz Dudek}
	\IEEEauthorblockA{\textit{Department of Electrical Engineering} \\
		\textit{Czestochowa University of Technology}\\
		Częstochowa, Poland \\
		grzegorz.dudek@pcz.pl}
}

\maketitle

\begin{abstract}
Feedforward neural networks are widely used as universal predictive models to fit data distribution. Common gradient-based learning, however, suffers from many drawbacks making the training process ineffective and time-consuming. Alternative randomized learning does not use gradients but selects hidden node parameters randomly. This makes the training process extremely fast. However, the problem in randomized learning is how to determine the random parameters. A recently proposed method uses autoencoders for unsupervised parameter learning. This method showed superior performance on classification tasks. In this work, we apply this method to regression problems, and, finding that it has some drawbacks, we show how to improve it. We propose  a  learning  method  of  autoencoders  that  controls  the produced  random  weights.  We  also  propose how to determine the biases of hidden nodes. We empirically compare autoencoder based learning with other randomized learning methods proposed recently for regression and find that despite the proposed improvement of the autoencoder based learning, it does not outperform its competitors in fitting accuracy. Moreover, the method is much more complex than its competitors.

\end{abstract}

\begin{IEEEkeywords}
autoencoder, feedforward neural networks, randomized learning algorithms
\end{IEEEkeywords}

\section{Introduction}

Feedforward neural networks (FNNs) have attracted a great deal of interest due to their excellence predictive performance and universal approximation capabilities. The most popular learning methods involve some kind of gradient descent algorithm to learn FNN weights iteratively. However, gradient-based methods suffer from many drawbacks making the learning process ineffective and time-consuming. This is because they are sensitive to the initial values of the parameters. The learning trajectory, starting with different initial parameters, leads to local minima of the loss function. Thus, globally optimal parameters are not guaranteed. Moreover, the learning process is time-consuming for complex target functions (TFs), big data sets and large FNN architectures. Randomized learning, such as a random vector functional link (RVFL) network \cite{Ige95}, has been proposed as an alternative to conventional FNNs iterative learning using gradients. In randomized learning, the parameters of the hidden nodes are selected randomly and stay fixed. They do not need to be tuned during the training stage. The only parameters that need to be learned are the output weights. This makes the optimization problem convex \cite{Pri15}. As such, it can be solved easily and quickly using a standard least-squares method. Despite this simplification, randomized FNN learning still possesses universal approximation capabilities, provided there are a sufficient number of nonlinear hidden nodes \cite{Ige95}, \cite{Hus99}. 

Many simulation studies reported in the literature show the high performance of the randomized FNN when compared to fully adaptable FNNs. Randomization, which is cheaper than optimization, ensures simplicity of implementation and faster training. However, a challenging and still open question in randomized learning is how to choose appropriate weights and biases for the hidden nodes to ensure best model performance \cite{Cao18}, \cite{Zha16}. 
To deal with this problem, various methods of generating hidden node parameters have been proposed. The simplest and most popular solution is to select both weights and biases from a uniform distribution over some symmetric interval $U=[-u, u]$. 
Usually, this interval is assigned as fixed, typically $[-1, 1]$, regardless of the data, TF, and type of activation functions (AFs). The independence of the hidden nodes from data is seen as an asset. This overly-simplistic approach was criticized as illogical and misleading \cite{Li17}. Therefore, to improve its performance,  optimization of interval $U$ for a specified application is recommended. For example, in \cite{Wan17},
a supervisory mechanism which randomly assigns hidden node parameters from an adaptively selected interval, was proposed.  This paper clearly reveals that the selection of random parameters should be data dependent to ensure the
universal approximation property of the resulting randomized FNN.

Data dependent random parameters were also recommended in \cite{Gor16}. The authors of this work noticed that if the hidden nodes are chosen at random and not subsequently trained, they are usually not placed in accordance with the density of the input data. In such a case, training of linear parameters is less ineffective at reducing errors. Therefore, in order to improve learning performance, the authors advise unsupervised placement of hidden nodes according to the input data density. This recommendation was implemented in the methods proposed in \cite{Dud19} and \cite{Dud20a}. In these works, it was noticed that as the weights and biases of hidden nodes have different functions, they should not be selected from the same interval. The weights decide about AF slopes and should reflect TF complexity, while the biases decide about the placement of AF in the input space. The biases should ensure the introduction of the most nonlinear fragments of AFs into the input hypercube. These fragments are most useful for modeling TF fluctuations. According to the methods described in \cite{Dud19} and \cite{Dud20a}, we first select the proper interval for the weights based on the AF features and TF properties. Then, the biases are calculated based on the weights and data distribution. In \cite{Dud20}, a data-driven method was proposed to improve further the FNN randomized learning. This method introduces the AFs into randomly selected regions of the input space and adjusts the slopes of individual AFs to the TF slopes in these regions. As a result, the AFs mimic the TF locally and their linear combination approximates smoothly the entire TF.  

An interesting method of generating parameters for RVFL was proposed recently in \cite{Tan21}. For this, the authors employ support-vector machines which in a supervised manner, by solving their corresponding optimization problems, generate the pre-trained weights. These weights are used to initialize the hidden layer of the proposed RVFL architecture.
An alternative approach to generating hidden nodes in FNNs is unsupervised parameter learning using autoencoders (AEs), which was first introduced in \cite{Kas13}. AE, in the encoding phase, transforms input data into a meaningful feature representation obtained from the hidden layer. Then, in the decoding phase, this feature representation is converted to the original inputs. The information hidden in original data can be explored and encoded into the output weights of AE \cite{Hin06}. These output weights are then introduced to FNN as hidden node weights instead of randomly generated weights \cite{Kas13}. This approach was applied in \cite{Zha19} for classification tasks. Here, RVFL uses a sparse AE with $\ell_1$-norm regularization to adaptively learn superior hidden node parameters for specific learning tasks. The authors claim that the learned network parameters in their sparse pre-trained RVFL are embedded with the valuable information about input data, which alleviates the randomly generated parameter issue and improves algorithmic performance.    

Another classifier based on RVFL with unsupervised parameter learning was proposed in \cite{Kat19}. In this solution, randomization based stacked AEs with a denoising criterion are used to extract better, higher-level representations. Each randomization based AE acts as an independent feature extractor and a
deep network is obtained by stacking several such AEs. The network is built hierarchically with high level
feature extraction followed by a final classification layer, which is RVFL with direct links. 

The authors of both works on AE based randomized learning, \cite{Zha19} and \cite{Kat19}, report experimental results on many real-world classification data sets from different domains. The results confirm the excellent effectiveness of the proposed solutions. Encouraged by these state-of-the-art results for FNN classifiers trained in an unsupervised manner using AEs, in this study, we analyze unsupervised parameter learning of FNN for regression. We compare this approach with alternative methods, which were proposed recently in \cite{Dud20a}.  

This paper makes the following contributions:
\begin{enumerate}
  \item We analyze AE based unsupervised parameter learning for a FNN regression model and find that this method has some drawbacks. We show how to improve it. We propose a learning method of AE that controls the produced random weights for FNN. We also propose how to determine the biases for FNN.   
  
  \item We empirically compare AE based learning with other randomized learning methods proposed recently for regression and find that AE based learning does not outperform its competitors in fitting accuracy, and, in fact, it is much more complex.

\end{enumerate}

The remainder of this paper is structured as follows. In Section II, we describe randomized FNN learning and methods for generating random parameters. AE based generating of random parameters is presented and critically analyzed from the perspective of AF distribution and shaping in Section III. In Section IV, we analyze the complexity of randomized AE. The performance of AE based randomized learning is evaluated in Section V. Finally, in Section VI, we conclude the work.  

\section{Randomized Learning of FNN}
\label{FNN}

Let us consider a shallow FNN architecture with $n$ inputs, a single-hidden layer including $m$ nonlinear nodes, and a single output. AFs of hidden nodes, $h_i(\mathbf{x})$, map nonlinearly input vectors $\mathbf{x}=[x_1, x_2,..., x_n]^T$ into $m$-dimensional feature space. An output node combines linearly $m$ nonlinear transformations of the inputs. FNN expresses a function in the form:   

\begin{equation}
\varphi(\mathbf{x}) = \sum_{i=1}^{m}\beta_ih_i(\mathbf{x})
\label{eq1}
\end{equation}
where $\beta_i$ is the output weight between the $i$-th hidden node and the output node.

Such shallow architecture has a universal approximation property, even when the hidden layer parameters are not trained but generated randomly from the proper distribution \cite{Ige95}, \cite{Hus99}. 

The output weights $ \boldsymbol{\beta} = [\beta_1, \beta_2, ..., \beta_m]^T$ can be determined by solving the following linear problem: $\mathbf{H}\boldsymbol{\beta} = \mathbf{Y}$, where $\mathbf{H} = [\mathbf{h}(\mathbf{x}_1), \mathbf{h}(\mathbf{x}_2),  ..., \mathbf{h}(\mathbf{x}_N)]^T \in \mathbb{R}^{N \times m}$ is the hidden layer output matrix, and $ \mathbf{Y} = [y_1, y_2,  ..., y_N]^T $ is a~vector of target outputs. Using Moore-Penrose pseudoinverse, the optimal solution is given by:

\begin{equation}
\boldsymbol{\beta} = \mathbf{H}^+\mathbf{Y}
\label{eq2}
\end{equation}
where $ \mathbf{H}^+ $ denotes the Moore–Penrose generalized inverse of matrix $ \mathbf{H} $.

The hidden node parameters, i.e. weights $ \mathbf{a} = [ a_{1}, a_{2}, ..., a_{n}]^T$ and bias $b$, control AF slope, orientation and position in the input space. For a sigmoid AF given by the formula: 

\begin{equation}
h(\mathbf{x}) = \frac{1}{1 + \exp\left(-\left(\mathbf{a}^T\mathbf{x} + b\right)\right)}
\label{eq3}
\end{equation}
weight $a_j$ expresses the sigmoid slope in the $j$-th direction and bias $b$ decides about the sigmoid shift along a hyperplane containing all $x$-axes. An appropriate selection of the slopes and shifts of all sigmoids determine the approximation properties of the model. 

As it was shown in  \cite{Dud19} and \cite{Dud20a}, the standard way of selecting both the hidden node weights and biases randomly, from the same interval, $a_{i,j}, b_i \sim U(-u, u)$, is misguided. This is because the optimal interval for weights, which determine the AF slope range, is not the optimal interval for biases, which represent an AF shift. And vice versa. With this in mind, in \cite{Dud19} and \cite{Dud20a} separate methods for determining weights and biases were proposed. The approach described in \cite{Dud19} first selects the weights $a_{i,j}$ from $U(-u, u)$. The bounds of the interval, $u$, are adjusted to TF complexity. For flat TFs we expect lower bounds, while for strongly fluctuating TFs we expect higher bounds. Once the weights have been selected, the biases are determined in such a way as to ensure the steepest fragments of the sigmoids (which are around their inflection points) are introduced into input hypercube $ H = [x_{1,\min}, x_{1,\max}]\times ... \times[x_{n,\min}, x_{n,\max}] $. The resulting equation for the $i$-th hidden node bias is as follows: 

\begin{equation}
b_i = -\mathbf{a}_i^T\mathbf{x}_i^*
\label{eqDer5a}
\end{equation}
where $\mathbf{x}_i^*=[x_{i,1}^*, ..., x_{i,n}^*]$ is a point from $H$ where the $i$-th sigmoid has one of its inflection points.   

As you can see from \eqref{eqDer5a}, the biases are dependent on the weights. Point $\mathbf{x}_i^*$ can be selected as follows:
\begin{itemize}
	\item this can be some point randomly selected from $H$: $\mathbf{x}_i^* \sim U(H)$. This method is suitable when the input points are evenly distributed in $ H $.
	\item this can be some randomly selected training point: $\mathbf{x}_i^* = \mathbf{x}_\xi \in \Phi $, where $\xi \sim U\{1, ..., N\}$, and $\Phi $ is a training set. This method distributes the sigmoids according to the data density, avoiding empty regions.
	\item this can be a prototype of the training point cluster: $\mathbf{x}_i^* = \mathbf{p}_i $, where $ \mathbf{p}_i $ is a~prototype of the $ i $-th cluster. This method requires the clustering of training points into $ m =$\#nodes clusters.   
\end{itemize}

It was noticed in \cite{Dud20a} that the relationship between weights $a$ and the slope angles of sigmoids $\alpha$ is highly nonlinear. The standard interval for $a$, $U=[-1, 1]$ corresponds to the interval $U_\alpha=[-14^\circ, 14^\circ]$ for $\alpha$, so only flat sigmoids are obtainable in such a case. To get steep sigmoids, with $\alpha$ near $90^\circ$, the bounds for $U$ should be $u>100$. For narrow $U$, such as $[-1, 1]$, the distribution of $\alpha$ is similar to a uniform one. When the interval for $a$ is extended, the $\alpha$ distribution changes such that larger angles, near the bounds of $U_\alpha$, are more probable than smaller ones. When $a \in [-100, 100]$, more than $77\%$ of sigmoids are inclined at an angle greater than $80^\circ$, so they are very steep. In such a case, there is a real threat of overfitting. To generate sigmoids with uniformly distributed slope angles, first, we select randomly $|\alpha_{i,j}| \sim U(\alpha_{\min}, \alpha_{\max})$, where the bound angles, $\alpha_{\min} \in (0^\circ, 90^\circ)$ and $\alpha_{\max} \in (\alpha_{\min}, 90^\circ)$, are adjusted to the TF complexity. Then, we calculate the weights from:

\begin{equation}
a_{i,j}=4 \tan \alpha_{i,j} 
\label{eq6a}
\end{equation}

Finally, to introduce the sigmoids into the input hypercube $H$, the biases are calculated from \eqref{eqDer5a}.

Both methods of generating random parameters of hidden nodes described above, we use in the experimental part of the work as comparative methods for AE based method. We denote them as R$a$M, i.e. a random weights $a$ method, and R$\alpha$M, i.e. a random slope angles $\alpha$ method. 

\begin{figure}[t]
	\centering
	\includegraphics[width=0.24\textwidth]{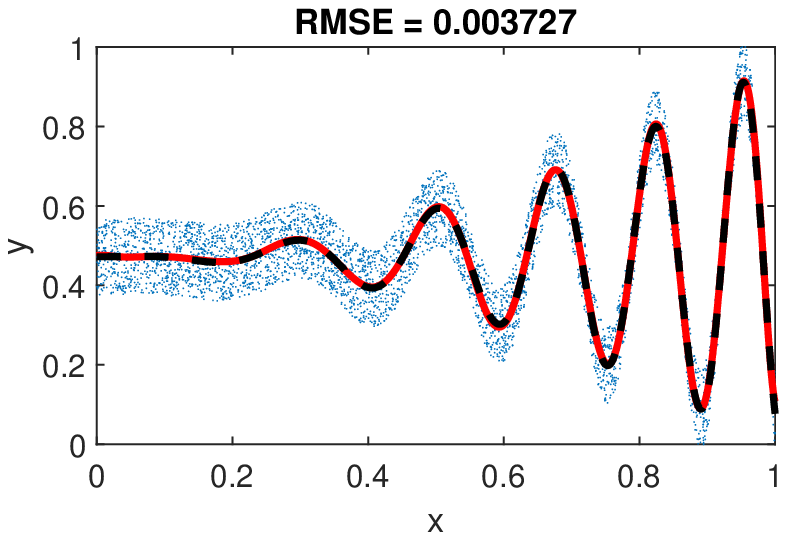}
	\includegraphics[width=0.24\textwidth]{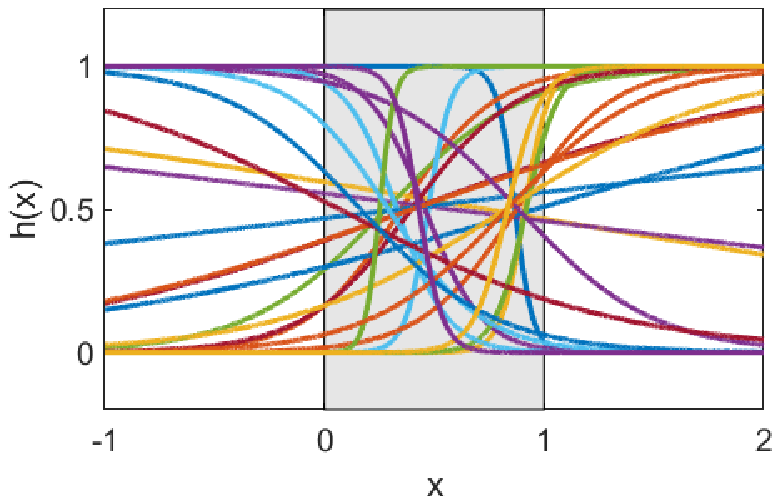}
	\caption{Fitted curve (left panel) and hidden node sigmoids (right panel) for R$\alpha$M with $|\alpha| \sim U(0^\circ, 83^\circ)$.}
	\label{figRalM}
\end{figure}

Fig. \ref{figRalM} illustrates an approximation of a highly nonlinear TF by FNN trained using R$\alpha$M (similar results were obtained for R$a$M). TF, shown as the dashed line in the left panel, is fitted accurately by the function built by FNN (red line). This fitted function is composed of 25 hidden node sigmoids, which are shown in the right panel. Note that the steepest fragments of the sigmoids are introduced by R$\alpha$M into the input interval, which is shown by the gray field in the panel on the right. This fragments are the most useful for modeling the TF fluctuations. The saturated AF fragments in the input interval are avoided. The interval for sigmoid slope angles, $U_\alpha =[0^\circ, 83^\circ]$, was adjusted to the TF complexity. This gives a perfect fitting.   

\section{Autoencoder based Generation of Hidden Nodes Parameters}
\label{AE}

Unsupervised parameter learning using AEs was proposed in \cite{Kas13}. In this work, randomization based autoencoders (RAE) are used for unsupervised feature extraction for a multilayer FNN classifier. RAE consists of two parts, an encoder and a decoder. The encoder maps the input data randomly into some latent representation in such a way that the decoder is able to reconstruct the original input data form this representation. RAE is a single hidden layer FNN with $n$ inputs, $n$ outputs, and $m$ nonlinear hidden nodes. The sigmoid AFs, $g(\mathbf{x})$, are used for the hidden layer and linear AFs are used for the output layer. The sigmoid AF is given by:

\begin{equation}
g(\mathbf{x}) = \frac{1}{1 + \exp\left(-\left(\mathbf{w}^T\mathbf{x} + c\right)\right)}
\label{eqg}
\end{equation}
where $\mathbf{w} = [ w_{1}, w_{2}, ..., w_{n}]^T $  are the hidden node weights and $ c$ is its bias.

In the first step, the learning method using RAE (RAEM) selects randomly 
the hidden node parameters for RAE, i.e. weights $\mathbf{w}_i$ and biases $ c_i, i = 1, 2, ..., m $. Typically, both are taken from the uniform distribution and interval $[-1,1]$. In the second step, the hidden layer output matrix is calculated: $\mathbf{G} = [\mathbf{g}(\mathbf{x}_1), \mathbf{g}(\mathbf{x}_2),  ..., \mathbf{g}(\mathbf{x}_N)]^T \in \mathbb{R}^{N \times m}$, where $\mathbf{g}(\mathbf{x}_l)= [g_1(\mathbf{x}_l), g_2(\mathbf{x}_l), ..., g_m(\mathbf{x}_l)] $ is a vector of hidden node outputs for input pattern $\mathbf{x}_l$. Finally, the output weight matrix, $\mathbf{V} = [ \mathbf{v}_1, \mathbf{v}_2, ..., \mathbf{v}_m]^T\in \mathbb{R}^{m \times n}$, where $ \mathbf{v}_i=[v_{i,1}, v_{i,2}, ..., v_{i,n}]$, is calculated from:

\begin{equation}
\mathbf{V} = \mathbf{G}^+\mathbf{X}
\label{eqv}
\end{equation}
where $ \mathbf{G}^+ $ denotes the Moore–Penrose generalized inverse of matrix $ \mathbf{G} $  and $\mathbf{X} \in \mathbb{R}^{N \times n} $ is an input matrix.

Due to randomized learning of RAE, the output weights, $\mathbf{V}$, can be obtained easily using a standard linear least-squares method. These weights are considered to be the latent features of input data \cite{Kas13}, and are used as the hidden node weights for FNN instead of random weights. Thus, $\mathbf{A} = \mathbf{V}^T$, i.e. $a_{j,i}=v_{i,j}$.       

As for the biases of hidden nodes $b_i$, it is hard to find in the literature, how they are selected in RAE based FNN learning. 
The exception is \cite{Zha19}, where the authors outline the way of which the biases were determined.
They calculate the bias for the $i$-th node as the mean value of this node weights: $b_i=1/n\sum_{j=1}^{n}a_{j,i}$.  Let us look at this case from the perspective of AF distribution in the input space.

When the bias is defined as the weight average, the inflection point of the sigmoid for the one-dimensional case, which is for $h(x)=0.5$, can be obtained from: 

\begin{equation}
h(x) = \frac{1}{1 + \exp\left(-\left(ax + b\right)\right)}=0.5 \rightarrow x = -\frac{b}{a} = -1
\label{eqh}
\end{equation}  
where we substituted $b=a$ assuming that a bias is the mean value of the weights.

\begin{figure}[t]
	\centering
	\includegraphics[width=0.24\textwidth]{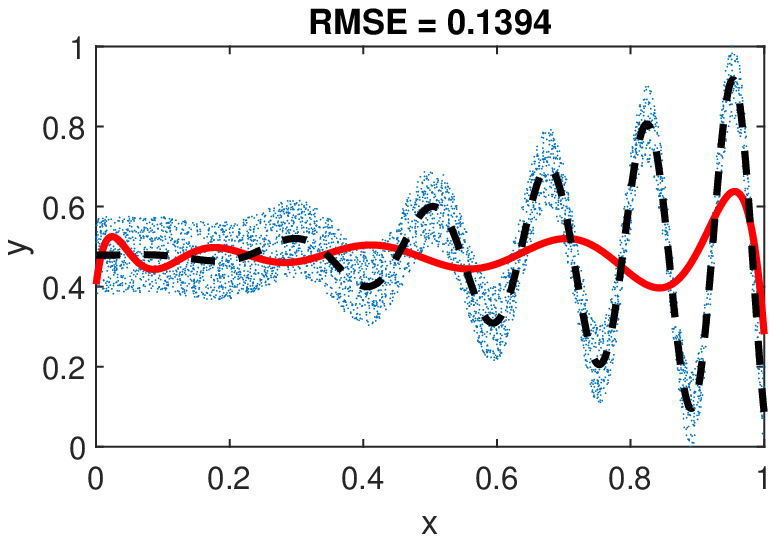}
\includegraphics[width=0.24\textwidth]{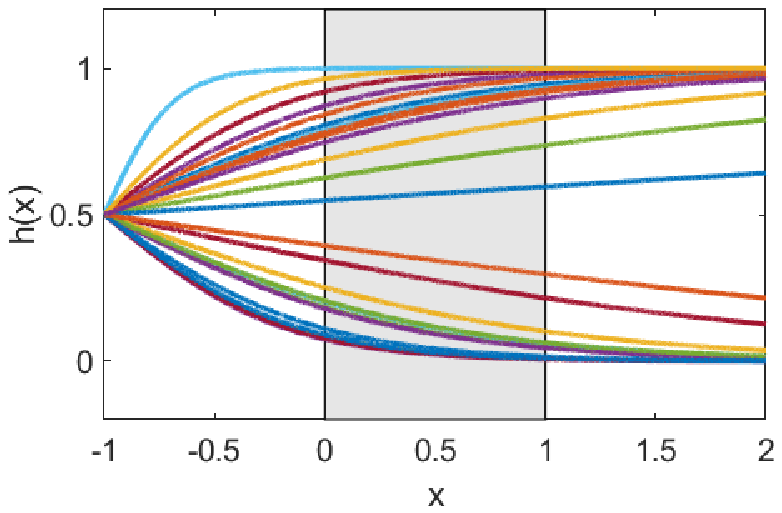}
	\caption{Fitted curve (left panel) and hidden node sigmoids (right panel) for RAEM with $b_i=a_i$.}
	\label{figFit}
\end{figure}

Thus, the inflection points of all sigmoids are in $x=-1$. This case is visualized in Fig. \ref{figFit}. Note that the steepest fragments of the sigmoids, which are around point $x=-1$, are far from the input interval. This interval includes many saturated fragments of the sigmoids, which results in poor fitting. The accuracy does not improve with the number of hidden nodes.

For the $n$-dimensional case, the sigmoid value in the inflection points is:  

\begin{equation}
h(\mathbf{x}) = \frac{1}{1 + \exp\left(-\left(\mathbf{a}^T\mathbf{x} + b\right)\right)}=0.5 \label{eq2}
\end{equation}
Substituting the mean value of weights for $b$ in this equation, after transformations we obtain:
\begin{equation}
\mathbf{a}^T\mathbf{x} + \overline{a} = 0\\
\label{eqp}
\end{equation}
where $\overline{a}=\frac{1}{n}\sum_{j=1}^{n}a_j$. 

This equation expresses the inflection hyperplane of the sigmoid. As we can see, this hyperplane is totally dependent on weights $a_j$. So, the weights generated by the RAE determine all sigmoid features, slopes in all directions and shift. The interdependence of the slopes and shift is an undoubted disadvantage of this approach. It is unjustified and limits the model's flexibility. The slopes should correspond to the TF complexity and the shift should be related to data distribution. Unfortunately, RAEM proposed in \cite{Zha19} cannot control separately the slopes and shifts of the sigmoids when it generates them. When ignoring biases and assuming $b_i=0$ for all hidden nodes, all sigmoids pass through $\mathbf{x} = (0, ..., 0)$. This is also an unacceptable solution as it limits the approximation properties of the model.

\begin{figure}[t]
	\centering
\includegraphics[width=0.24\textwidth]{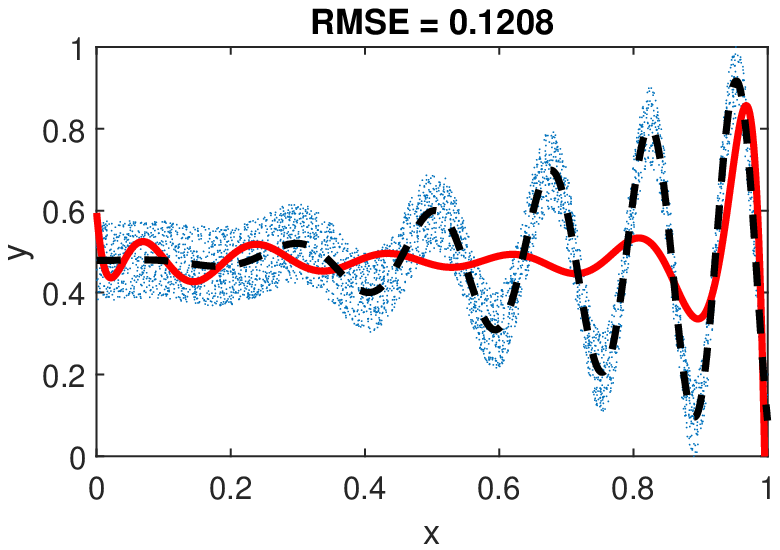}
\includegraphics[width=0.24\textwidth]{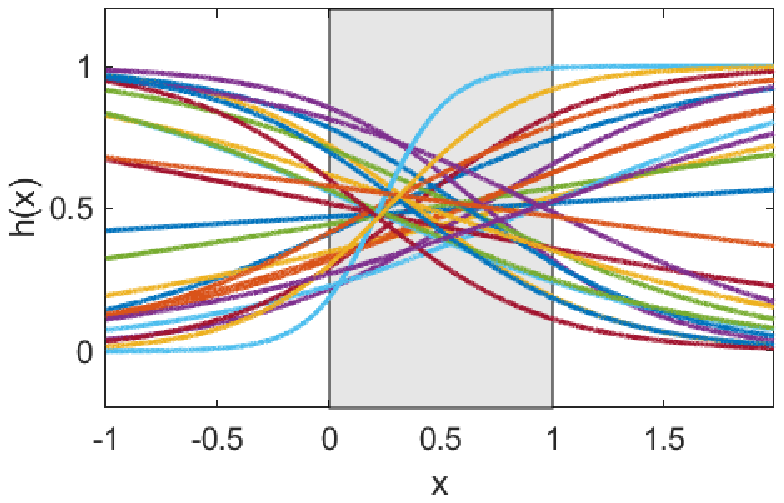}
	\caption{Fitted curve (left panel) and hidden node sigmoids (right panel) for RAEM with $b_i=-a_ix^*_i$.}
	\label{figFit2}
\end{figure}

In an attempt to improve the RAEM performance, we use the same solution for biases as in R$a$M and R$\alpha$M. We calculate them from \eqref{eqDer5a} on the basis of weights $\mathbf{a}_i$ produced by RAE and randomly selected points $\mathbf{x}^*$. Thus, we distribute the sigmoids in $H$ according to the data distribution. The result for the one-dimensional case is shown in Fig. \ref{figFit2}. As we can see, the accuracy did not improve significantly. Moreover, when we add new hidden nodes (up to 2000), the RMSE remains unacceptable, i.e. above 0.1. This means that the problem is not only due to the mis-determination of the biases, but also due to the too flat sigmoids that do not correspond to TF complexity. 

The sigmoid slopes in FNN are determined by the RAE output weights, $v$. These weights are dependent on the RAE random projection $\mathbf{G}$, which in turn is dependent on the RAE hidden node random parameters: $w$ and $c$. In the above described simulations, these parameters were both selected from the standard interval $U_{AE}=[-1, 1]$. As we shown in Section \ref{FNN}, this is an incorrect approach. Thus, to improve the regression model performance, we propose to use R$a$M for generating $w$ and $c$. According to this method, we optimize the interval for $w$, $U_{AE}=[-u_{AE}, u_{AE}]$, and calculate the biases $c$ analogously to \eqref{eqDer5a}: 

\begin{equation}
c = -\mathbf{w}^T\mathbf{x}^*
\label{eqc}
\end{equation}  

As a result of this modified RAEM learning, we expect that RAE will produce the appropriate weights $\mathbf{A} = \mathbf{V}^T$, which provide FNN sigmoids with the slopes adjusted to the TF complexity. 

\begin{figure}[t]
	\centering
	\includegraphics[width=0.24\textwidth]{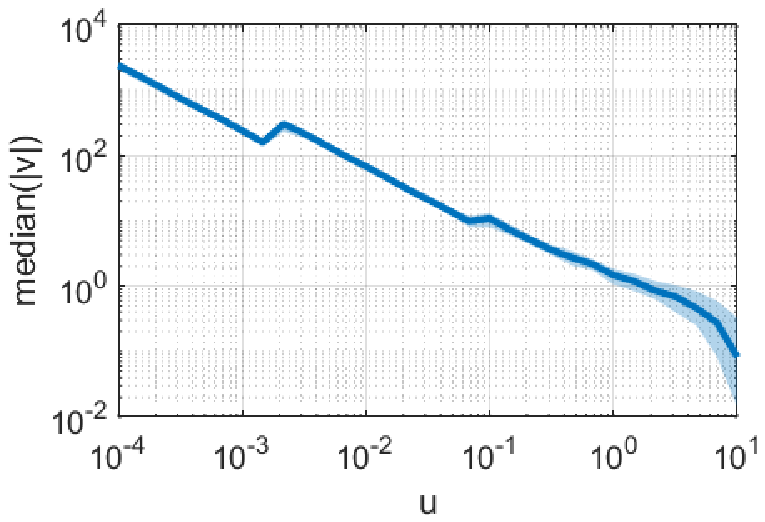}
	\includegraphics[width=0.24\textwidth]{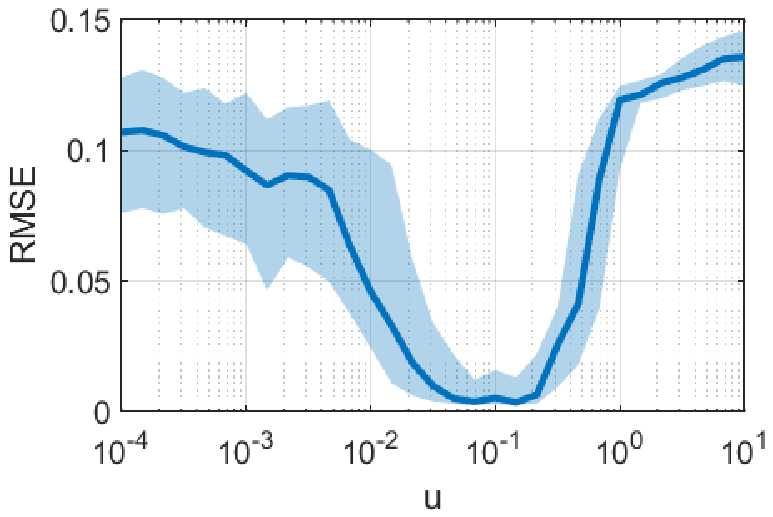}
	\caption {The effect of the interval $U_{AE}$ bounds on the median of absolute values of weights $v$ (left panel) and on the fitting error (right panel).}
	\label{figU}
\end{figure}

Fig. \ref{figU} shows the effect of the interval $U_{AE}$ bounds on the median of absolute values of weights $v$ (left panel) and on the fitting error (right panel) for TF shown in Figs. \ref{figRalM}-\ref{figFit2} (the number of hidden nodes was  $m=25$, results are averaged over 100 runs). Note that median of $|v|$ decreases quickly with $u_{AE}$. RMSE reaches its minimum for $u_{AE} \in(0.05, 0.2)$. For such $U_{AE}$ bounds, RAE produces weights $v$  whose $median|v|$  is in the range from around 5 to 14 (for comparison, when weights $w$ were selected from the standard interval $U_{AE}=[-1, 1]$,  $median|v|$ was around 1.5). For such a case, RMSE reaches an acceptable level of 0.005, which is similar to those obtained by R$a$M and R$\alpha$M for the same number of hidden nodes. 

\begin{figure}[t]
	\centering
	\includegraphics[width=0.24\textwidth]{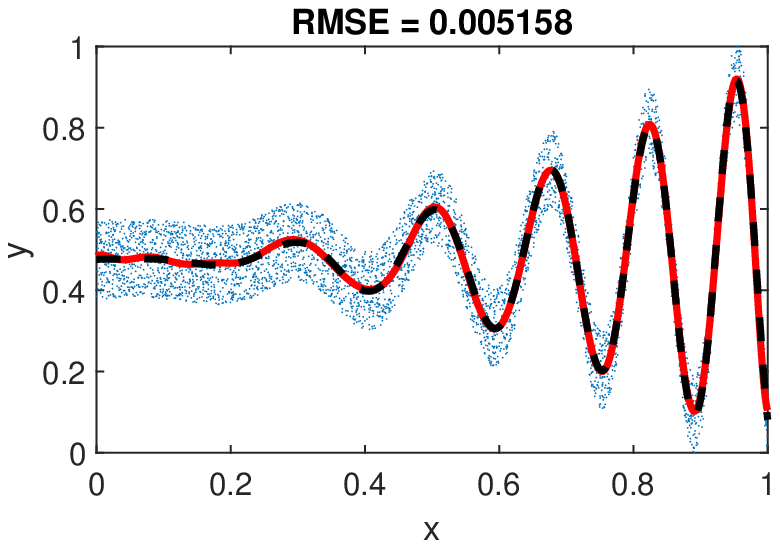}
	\includegraphics[width=0.24\textwidth]{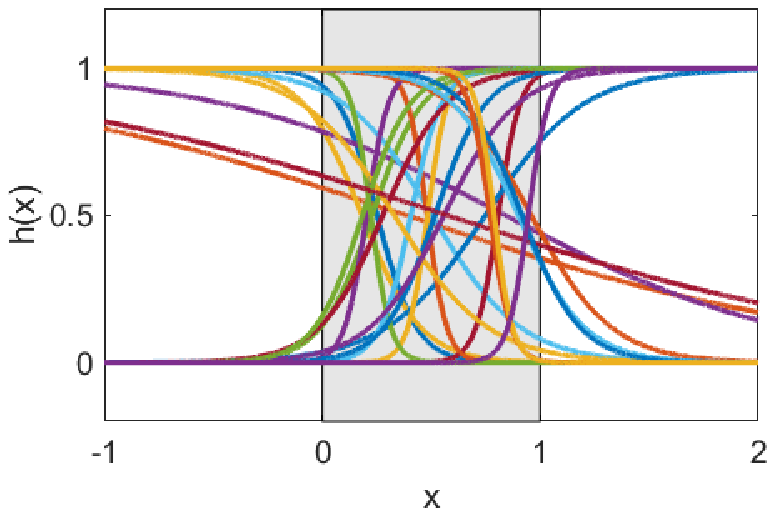}
	\caption{Fitted curve (left panel) and hidden node sigmoids (right panel) for RAEM with $w_i \in [-0.1, 0.1]$ and $c_i=-w_ix^*_i$.}
	\label{figFitU}
\end{figure}

Fig. \ref{figFitU} depicts fitting results for $U_{AE}=[-0.1, 0.1]$. Note steeper sigmoids than in the cases presented in Fig. \ref{figFit2}, resulting in good fitting. It should be noted that the optimal interval $U_{AE}$ depends on the number of hidden nodes. When, instead of 25, we used 200 hidden nodes, the optimal values of the interval bounds were $u_{AE} \in(0.005, 0.022)$. In such a case $median|v|$ was from around 4 to 17. Compare these values with $median(|v|) \approx 0.19$ which we obtain for $U_{AE}=[-1, 1]$. 

In the above analysis, RAE was trained without regularization. However, $\ell_1$, $\ell_2$ and elastic-net regularization are widely used in RAE to prevent overfitting \cite{Kas13}, \cite{Zha19}, \cite{Kat19}. Regularization in RAE decreases weights $v$ to improve the generalization property of the model, i.e. generalization of mapping $\mathbf{x}$ to themselves. From the point of view of the regression FNN model trained using RAEM,  regularisation in RAE is an unfavorable operation because it further flattens FNN sigmoids. As we showed above, it is a disadvantage for strongly fluctuating TFs. 

Note that RAEM in its standard version (without optimization of $U_{AE}$) produces weights $v$ dependent only on the distribution of points $\mathbf{x}$ in the input space and independently of the TF. So, for two TFs with different complexity (one of them flat and the second one with strong fluctuations), having the same distributed x-points, RAE can produce exactly the same weights $v$. This must be considered a serious drawback.

These findings on RAEM can be summarised in tree points:

\begin{itemize}
	\item the output weights $v$ produced by RAE determine the sigmoid slopes in the FNN regression model and are crucial for accurate TF approximation. These weights are dependent on the RAE hidden nodes parameters, i.e. weights $w$ and biases $c$. The standard way of generating both these parameters from the same interval  $U_{AE}$  is unjustified and misleading. We recommend optimizing this interval for weights $w$ and determining biases $c$ from \eqref{eqc}.
    \item the optimal interval  $U_{AE}$  for $w$ depends on the number of hidden nodes $m$. Thus, for each value of $m$ considered, interval $U_{AE}$ should be optimized.
   \item the hidden nodes biases of the FNN regression model, $b$, determine the sigmoid placement in the input space. They should introduce the steepest fragments of the sigmoids into the input hypercube. Incorrectly selected, such as $b_i=\overline{a}_i$ or $b_i=0$, they lead to the placement of the saturation parts of the sigmoids into $H$. To avoid this, we recommend determining biases $b$ from \eqref{eqDer5a}.
\end{itemize}




\section{Complexity of RAE}
\label{CC}

To generate weights $v$, RAE can use linear least squares regression, lasso, ridge regression or elastic net algorithms \cite{Kat19}. In all these cases the total time complexity is $O(Nm^2+m^3)$ (this is true for lasso when least-angle regression (LARS) is used for fitting linear regression models \cite{Efr04}). For $N>m$ the RAE runs in linear time with the size of the training set, $N$, and in quadratic time with the number of nodes, $m$. For $N\leq m$, it runs in cubic time with $m$, although in \cite{Has09} it was shown that for ridge regression this cost can be reduced to $O(N^2m)$.


Taking into account the optimization process, i.e. the selection of hyperparameters, the time complexity is as follows. In RAE without regularization the number of hidden nodes $m$ and interval $U_{AE}$ need to be found using cross-validation. Thus, the computational load increases linearly with the number of data splits used in cross-validation, $s$, and also with the number of points of the grid which is $l_ml_u$, where $l_m$ and $l_u$ are the number of searching values for $m$ and $u_{AE}$, respectively. So, the total complexity will be $O(sl_ml_uNm^2+sl_ml_um^3)$. In RAE with $\ell_1$ and $\ell_2$  regularization, the regularization parameter $\lambda$ should also be selected in cross-validation. So, the complexity of RAE with lasso or ridge regression is $O(sl_ml_ul_{\lambda}Nm^2+sl_ml_ul_{\lambda}m^3)$, where $l_{\lambda}$ is the number of searching points for $\lambda$. 
Elastic net combines ridge and lasso, with the tuning parameter $\alpha$ that balances the weights of ridge against lasso. If the number of searching points for $\alpha$ is $l_{\alpha}$, the total complexity of RAE with elastic net regularization is $O(sl_ml_ul_{\lambda}l_{\alpha}Nm^2+sl_ml_ul_{\lambda}l_{\alpha}m^3)$. 


\section{Simulation Study}
\label{SS}




In this section, to demonstrate the fitting properties of RAEM, we report some simulation results over several regression problems. They include an approximation of extremely nonlinear TFs: 

\begin{description}
\item[TF1] $g(\mathbf{x}) = \sum_{j=1}^{n}\sin\left(20\cdot\exp x_j\right)\cdot x_j^2, \, x_i \in [0, 1]$ 

\item[TF2] $g(\mathbf{x}) = -{\sum_{i=1}^{n} \sin(x_i) \sin^{20} \left(\frac{ix_i^2}{\pi}   \right)}, \, x_i \in [0, \pi]$ 

\item[TF3] $g(\mathbf{x}) = 418.9829n -{\sum_{i=1}^{n} x_i \sin(\sqrt{|x_i|})}, \\ x_i \in [-500, 500]$

\end{description}

\begin{figure}[]
	\centering
	\includegraphics[width=0.158\textwidth]{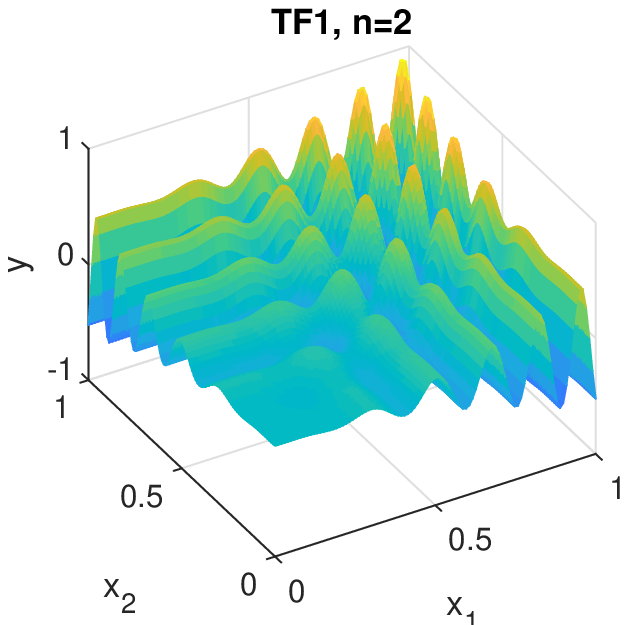}
	\includegraphics[width=0.158\textwidth]{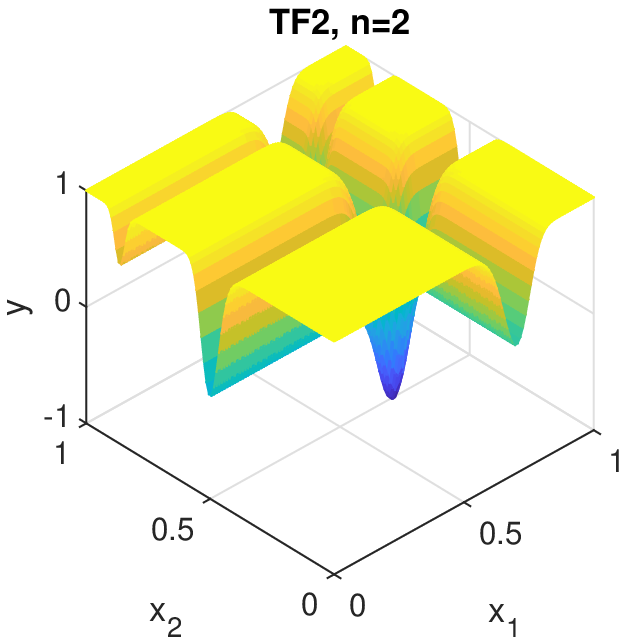}
	\includegraphics[width=0.158\textwidth]{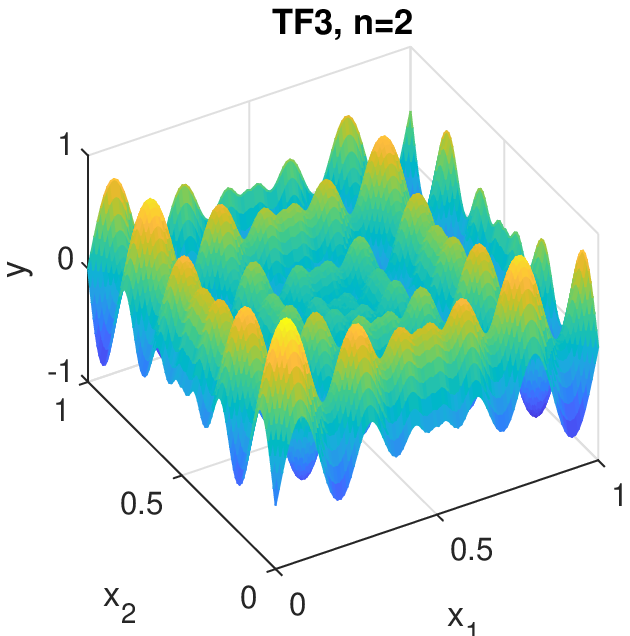}
	\caption{Target functions TF1-TF3 for $n=2$.} 
	\label{figTF35}
\end{figure}

We considered these functions with $n=2, 5$ and 10 arguments. The sizes of the training and test sets depended on the number of arguments. They were 5000 for $n=2$, 20,000 for $n=5$, and 50,000 for $n=10$. All arguments for TF3-TF5 were normalized to $[0, 1]$, and the function values were normalized to $[-1, 1]$. Two-argument functions TF1-TF3 are shown in Fig. \ref{figTF35}. Note that TF1 combines flat regions with strongly fluctuating regions, TF2 expresses flat regions with perpendicular grooves, and TF3 fluctuates strongly, showing the greatest amplitude at the borders.      

We use a modified version of RAEM. That is, the interval for the hidden nodes in RAE, $U_{AE}$, was optimized for each number of hidden nodes. This is because, as we show in Section \ref{AE}, the optimal interval is dependent on the hidden node number. To introduce the sigmoids of RAE into input hypercube $H$, biases $c$ were calculated from \eqref{eqc}. And analogously, to introduce the sigmoids of the FNN regression model into $H$, we calculate biases $b$ from \eqref{eqDer5a}. For comparison, we use R$a$M and R$\alpha$M. For each experiment, we run 100 independent training sessions.  

In Table \ref{tab1}, we present the fitting test errors (RMSE) of each method for each TF. The optimal bounds of the intervals from which the model parameters ($w$, $a$ and $\alpha$, respectively) were randomly selected are also shown. In R$\alpha$M, we set fixed lower bounds, $\alpha_{\min}=0^\circ$, while the upper bounds, $\alpha_{\max}$, were selected for each TF as $90^\circ$. For RAEM and R$a$M, we observe the difference in the optimal interval sizes between  2-argument and more than 2-argument TFs. R$a$M provides wide intervals, i.e. steep sigmoids, for 2-argument TFs, and narrow intervals, i.e. flat sigmoids, for 5- and 10-argument TFs. Similarly RAEM provides steep sigmoids for 2-argument TFs (narrow $U_{AE}$ produces higher weights $a$), and flat sigmoids for 5- and 10-argument TFs. This could be explained by the change in the TF landscape, which flattens with an increasing number of dimensions. Interestingly, R$\alpha$M is insensitive to this phenomenon, giving the same broad interval for $\alpha$ regardless of the number of arguments. 

\begin{figure}[t]
	\centering
\includegraphics[width=0.158\textwidth]{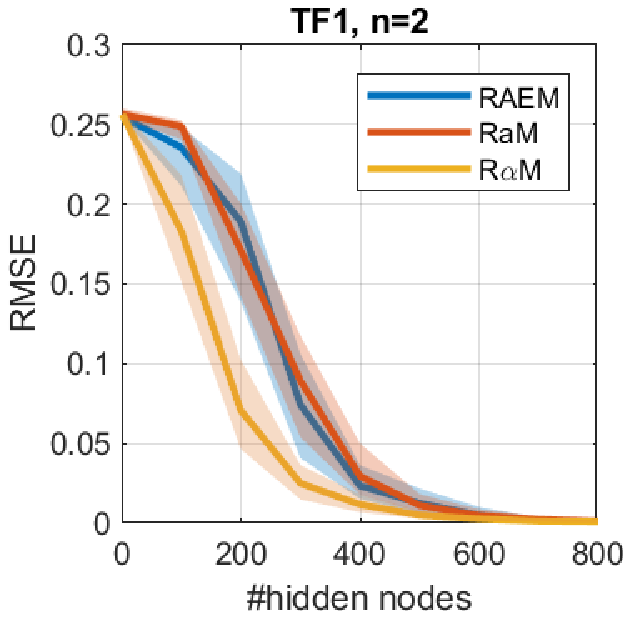}
\includegraphics[width=0.158\textwidth]{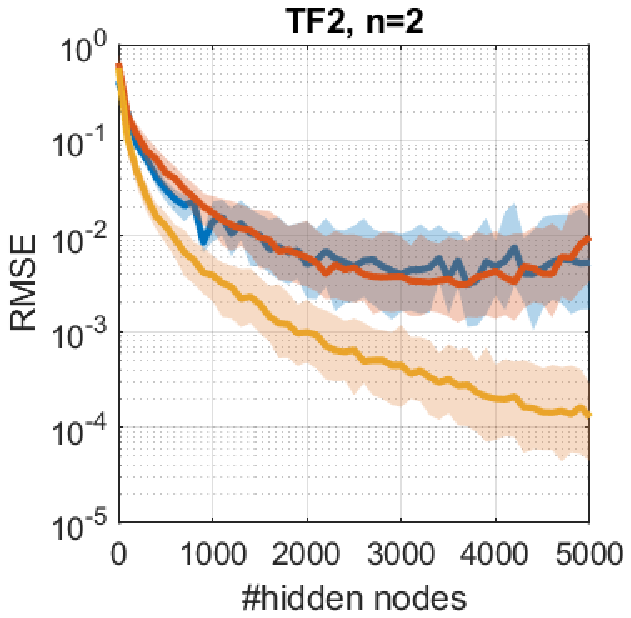}
\includegraphics[width=0.158\textwidth]{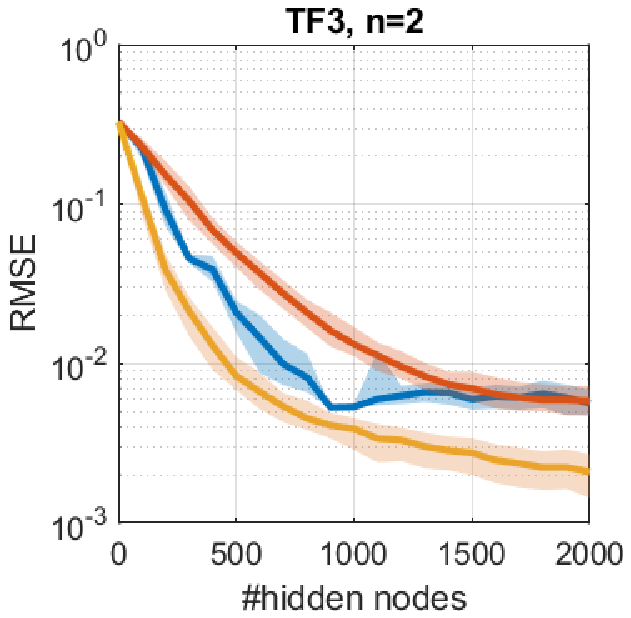}
\caption{RMSE depending on the hidden node numbers for 2-argument TFs.}
	\label{figZb1}
\end{figure}

\begin{figure}[t]
	\centering
\includegraphics[width=0.158\textwidth]{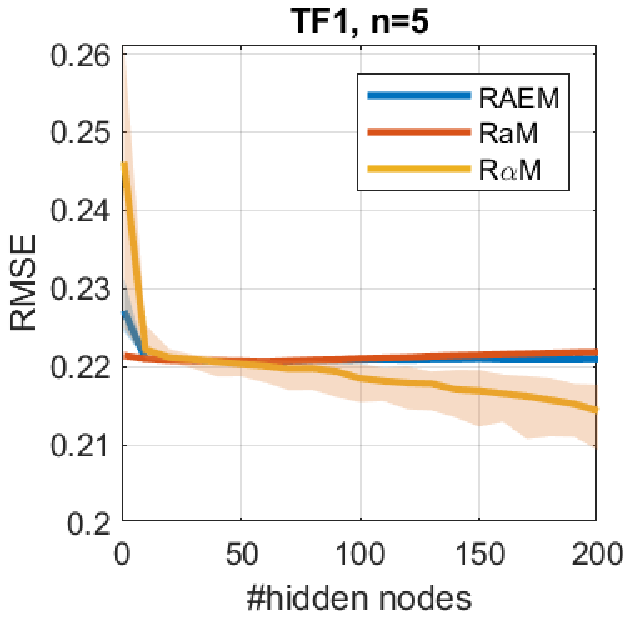}
\includegraphics[width=0.158\textwidth]{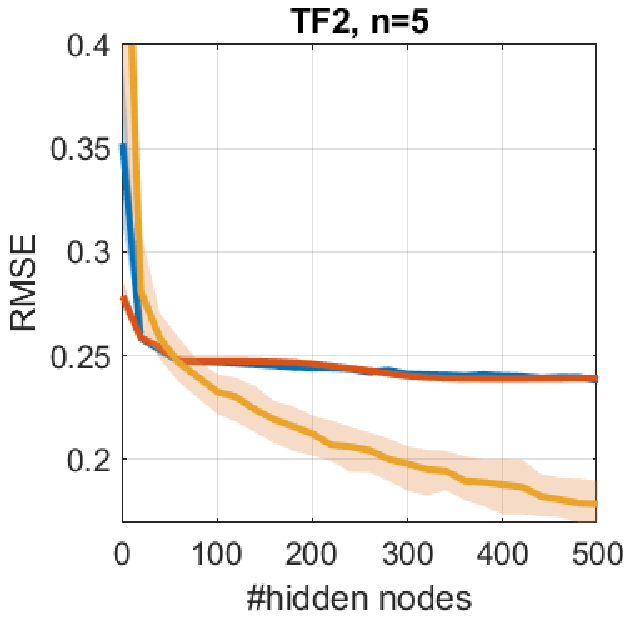}
\includegraphics[width=0.158\textwidth]{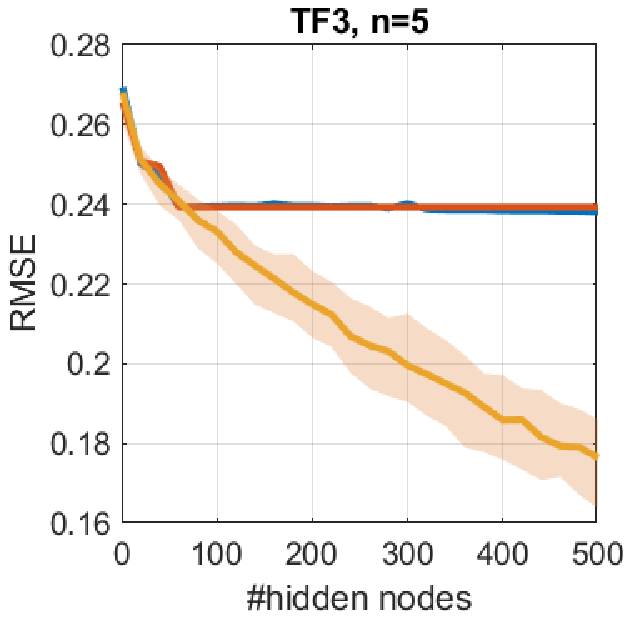}
	\caption{RMSE depending on the hidden node numbers for 5-argument TFs.}
	\label{figZb2}
\end{figure}

\begin{figure}[t]
	\centering
\includegraphics[width=0.158\textwidth]{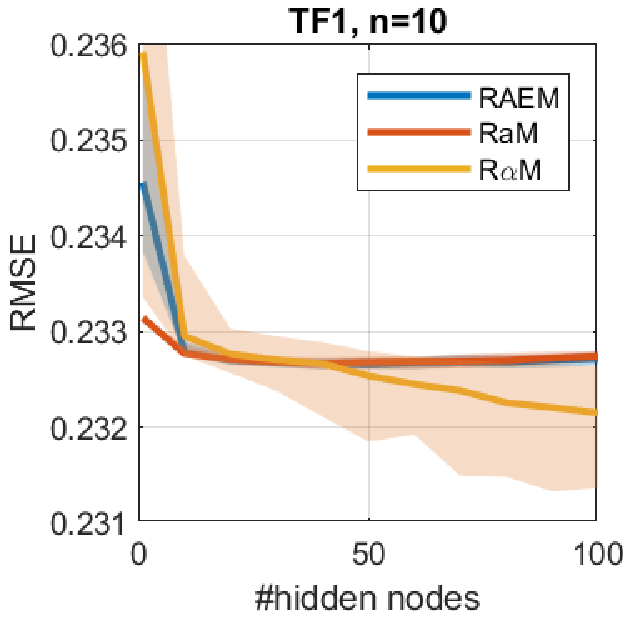}
\includegraphics[width=0.158\textwidth]{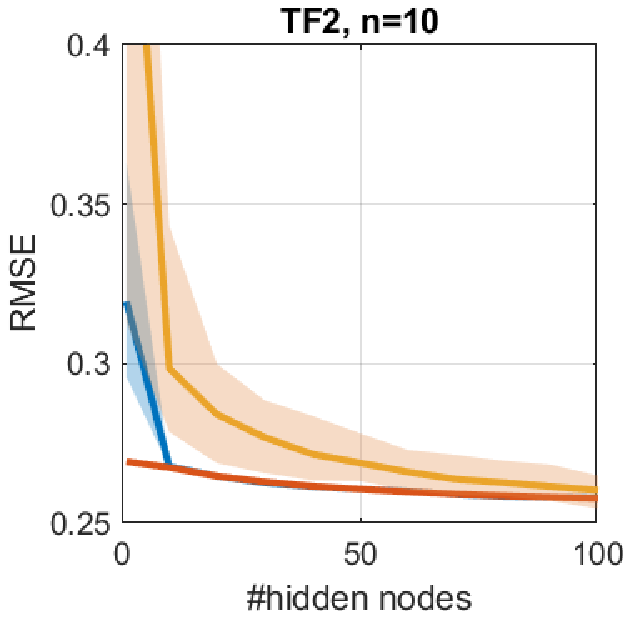}
\includegraphics[width=0.158\textwidth]{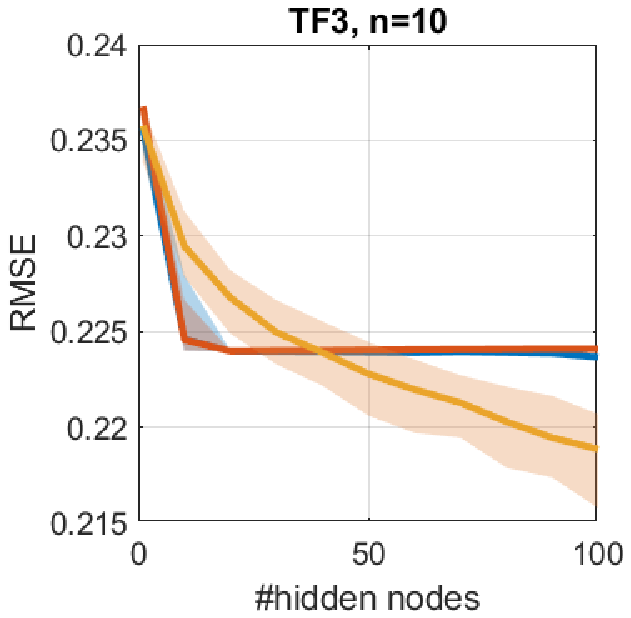}
	\caption{RMSE depending on the hidden node numbers for 10-argument TFs.}
	\label{figZb3}
\end{figure}

\begin{figure}[t]
	\centering
\includegraphics[width=0.158\textwidth]{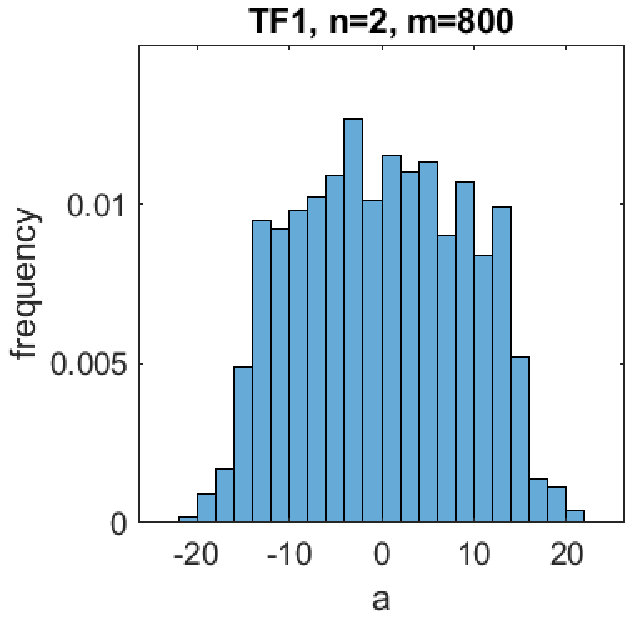}
\includegraphics[width=0.158\textwidth]{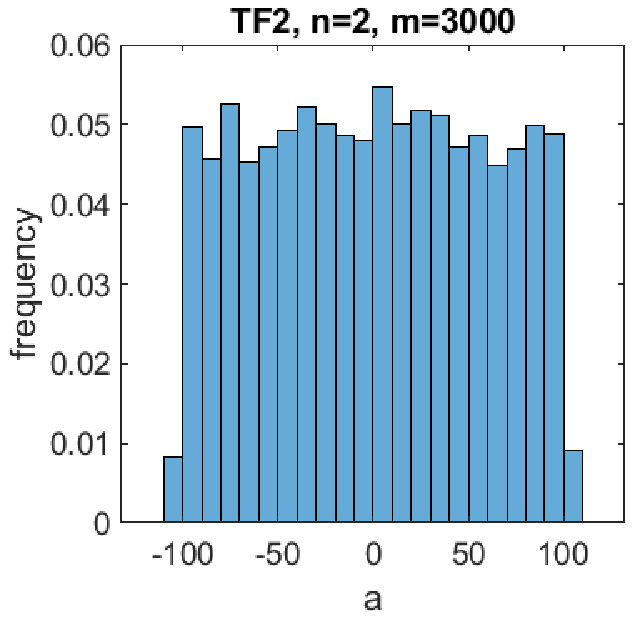}
\includegraphics[width=0.158\textwidth]{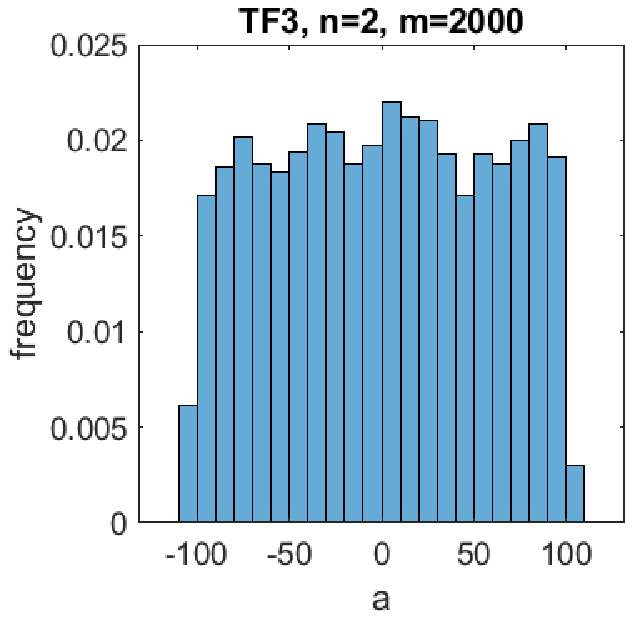}
	\caption{Histograms of weights $a$ generated by RAEM for 2-argument TFs.}
	\label{figh1}
\end{figure}

\begin{figure}[t]
	\centering
\includegraphics[width=0.158\textwidth]{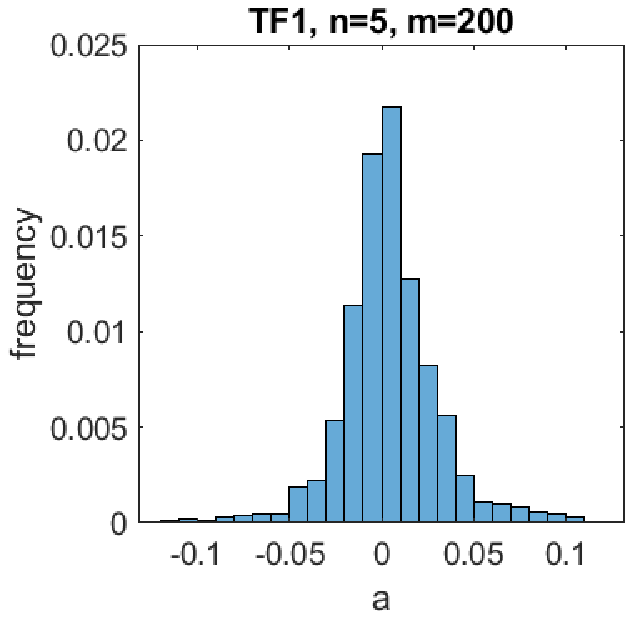}
\includegraphics[width=0.158\textwidth]{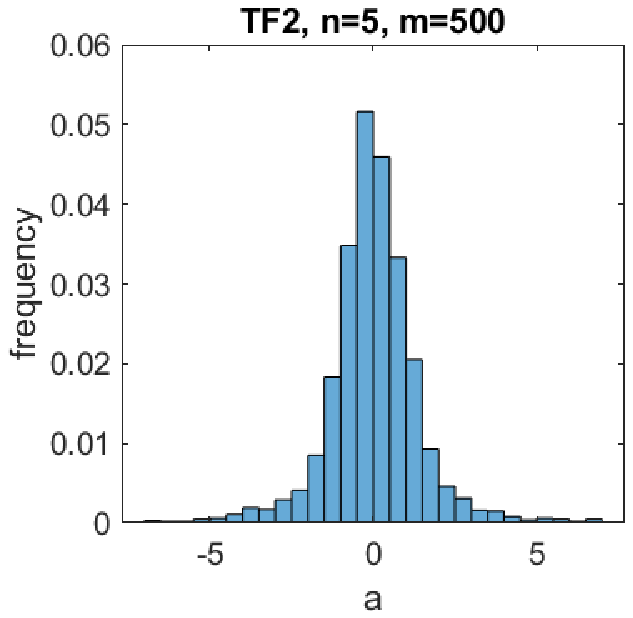}
\includegraphics[width=0.158\textwidth]{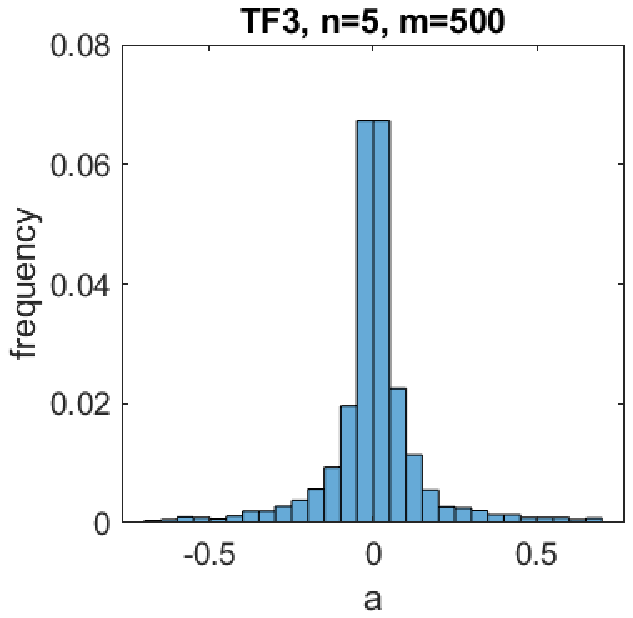}
	\caption{Histograms of weights $a$ generated by RAEM for 5-argument TFs.}
	\label{figh2}
\end{figure}

\begin{figure}[t]
	\centering
\includegraphics[width=0.158\textwidth]{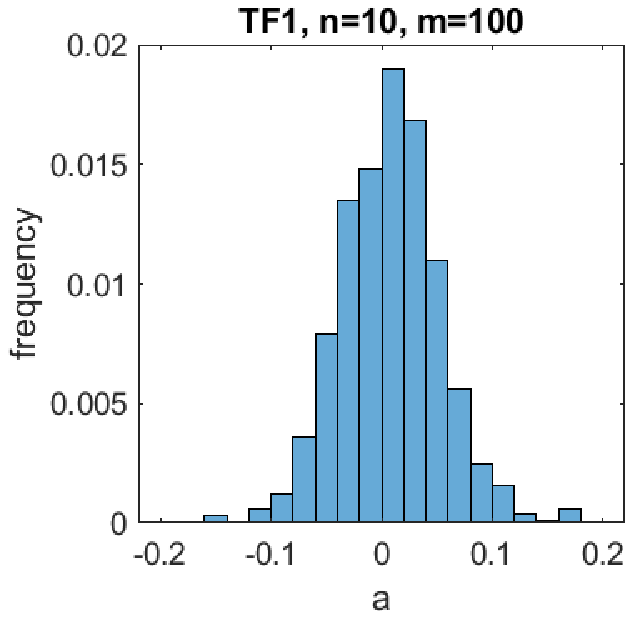}
\includegraphics[width=0.158\textwidth]{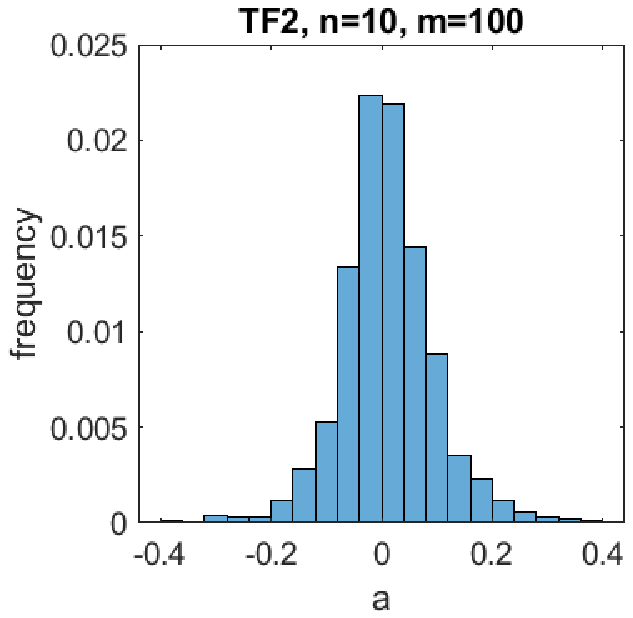}
\includegraphics[width=0.158\textwidth]{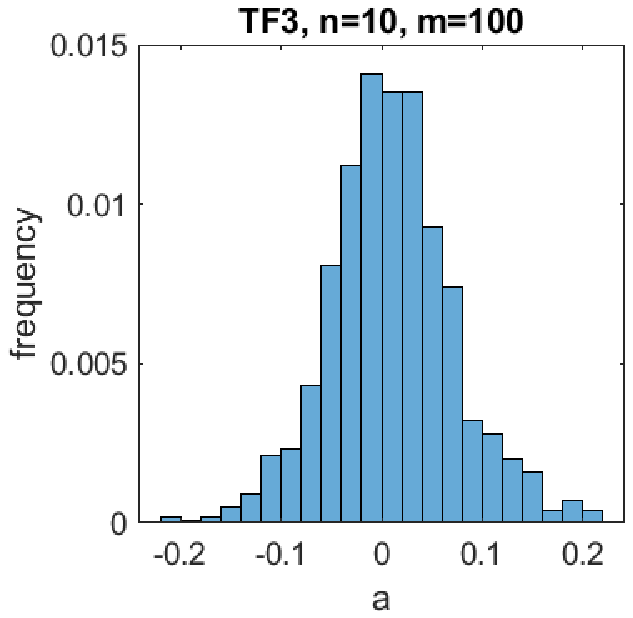}
	\caption{Histograms of weights $a$ generated by RAEM for 10-argument TFs.}
	\label{figh3}
\end{figure}

As can be seen from  Table \ref{tab1}, R$\alpha$M demonstrates the highest fitting accuracy for all TFs. This was confirmed by a Wilcoxon signed-rank test with $\alpha=0.05$. Note that results for RAEM and R$a$M are very similar.
Figs. \ref{figZb1}-\ref{figZb3} depict fitting test errors depending on the number of hidden nodes $m$. Shaded regions are 10th and 90th percentiles, measured over 100 trials. As can be seen from these figures, the confidence intervals of RAEM and R$a$M overlap for higher $m$. This means that these two methods generate similar weights $a$, which provide a similar set of basis functions for FNN. To study this issue further, we show in Figs. \ref{figh1}-\ref{figh3} the histograms of weights $a$ generated by RAEM for all TFs. It is evident from these figures that weight $a$ distributions deviate from the uniform distribution, especially for $n=5$ and $10$. Thus, RAEM, unlike R$a$M, does not generate uniformly distributed weights $a$. The weight distributions are unimodal, symmetrical, bell-shaped, and centered at 0. Note that the intervals for $a$ observed in Figs. \ref{figh1}-\ref{figh3} in most cases correspond to the intervals $U$ selected as optimal by R$a$M (see hyperparameter $u$ for R$a$M in Table \ref{tab1}). 

\begin{table*}[]
\begin{center}
\caption{Results for TF1-TF3.}
\label{tab1}
\begin{tabular}{lc|cc|cc|cc}
\toprule
           & \#nodes & \multicolumn{2}{c|}{RAEM}                               & \multicolumn{2}{c|}{R$a$M}                 & \multicolumn{2}{c}{R$\alpha$M}                            \\
           & $m$     & RMSE                                & $u_{AE}$         & RMSE                                & $u$  & RMSE                                & $\alpha_{\max}$ \\
\midrule
TF1, $n=2$  & 800                         & $0.0011 \pm 0.00050$    & 0.00100                      & $0.0012 \pm 0.00093$            & 20                      & $\mathbf{0.0006 \pm 0.00040}$     & 90                                 \\
TF2, $n=2$  & 3000                        & $0.0051 \pm 0.00407$    & 0.00004                      & $0.0049 \pm 0.00338$            & 100                     & $\mathbf{0.0006 \pm 0.00051}$     & 90                                 \\
TF3, $n=2$  & 2000                        & $0.0058 \pm 0.00095$    & 0.00006                      & $0.0059 \pm 0.00097$            & 100                     & $\mathbf{0.0024 \pm 0.00062}$     & 90                                 \\
TF1, $n=5$  & 200                         & $0.2210 \pm 0.00012$     & 8                            & $0.2219 \pm 0.00010$            & 0.1                     & $\mathbf{0.2137 \pm 0.00374}$      & 90                                 \\
TF2, $n=5$  & 500                         & $0.2381 \pm 0.00101$     & 0.1                          & $0.2392 \pm 0.00012$            & 0.5                     & $\mathbf{0.1734 \pm 0.00912}$      & 90                                 \\
TF3, $n=5$  & 500                         & $0.2381 \pm 0.00051$     & 1                            & $0.2392 \pm 5 \cdot 10^{-8}$ & 0.02                    & $\mathbf{0.1717 \pm 0.00903}$      & 90                                 \\
TF1, $n=10$ & 100                         & $0.2327 \pm 0.00006$     & 6                            & $0.2327 \pm 0.00005$            & 0.01                    & $\mathbf{0.2318 \pm 0.00072}$      & 90                                 \\
TF2, $n=10$ & 100                         & $0.2579 \pm 0.00043$     & 2                            & $0.2577 \pm 0.00034$            & 0.1                     & $\mathbf{0.2551 \pm 0.00290}$      & 90                                 \\
TF3, $n=10$ & 100                         & $0.2237 \pm 0.00013$     & 3                            & $0.2241 \pm 0.00006$            & 0.01                    & $\mathbf{0.2188 \pm 0.00212}$      & 90                                
   \\
\bottomrule
\end{tabular}
\end{center}
\end{table*}

\begin{figure}[t]
	\centering
\includegraphics[width=0.158\textwidth]{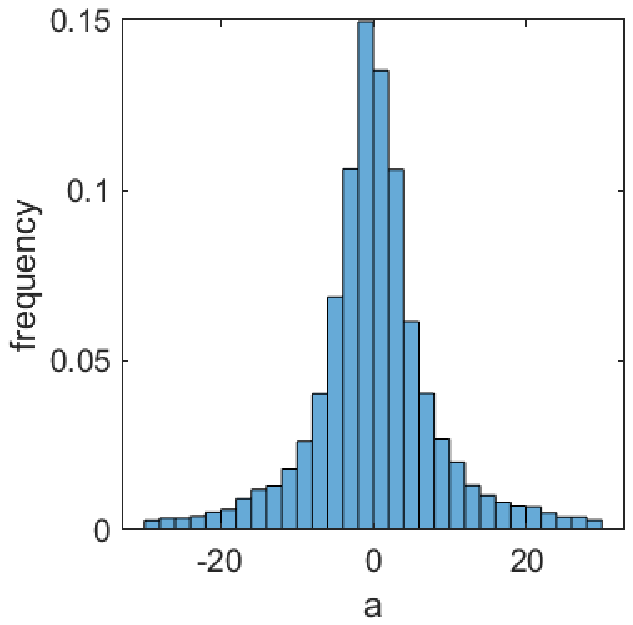}

	\caption{Histograms of weights $a$ generated by R$\alpha$M for all TFs.}
	\label{figh}
\end{figure}

It is clear from Table \ref{tab1} and Figs. \ref{figZb1}-\ref{figZb3} that the best fitting was achieved for R$\alpha$M. Interestingly, for each TF this method generated weights $a$ from the same distribution, which is shown in Fig. \ref{figh}.     

To compare the performance of RAEM, R$a$M and R$\alpha$M on real-world regression problems, we performed experiments on several data sets from different domains. Data sets were collected from the KEEL repository, http://www.keel.es/ (stock, laser, treasury, dee, and machineCPU data sets) and from Delve repository,  https://www.cs.toronto.edu/$\sim$delve/data/kin/desc.html (kin8nm data set).    
The number of samples and arguments in each data set are shown in Table \ref{tab2}. The input and output variables were normalized into $[0, 1]$. The data sets were divided into training sets containing 75\% of the samples selected randomly, and test sets containing the remaining samples. The optimal values of hyperparameters, i.e. hidden node numbers and sizes of intervals for random parameters, were selected by 5-fold cross-validation. Results are shown in Table \ref{tab2}. The bold values indicate the lowest errors while the values in italics indicate the highest errors (Wilcoxon signed-rank test with $\alpha=0.05$ was used to compare errors). Note that RAEM demonstrate the highest errors for four data sets. The best method is R$\alpha$M, which yielded the lowest errors for five out of six data sets.

\begin{table*}[]
\begin{center}
\caption{Results for real-word regression problems.}
\label{tab2}
\begin{tabular}{lc|ccc|ccc|ccc}
\toprule
Data set & \#samples/ & \multicolumn{3}{c|}{RAEM}              & \multicolumn{3}{c|}{R$a$M}        & \multicolumn{3}{c}{R$\alpha$M}              \\
       & \#arguments    & RMSE                & $m$  & $u_{AE}$ & RMSE                & $m$  & $u$ & RMSE                & $m$  & $\alpha_{\max}$ \\
\midrule
stock      & 950/9 & $\mathit{0.0303   \pm 0.0013}$ & 150  & 0.1585 & $0.0296 \pm 0.0016$          & 200  & 4   & $\mathbf{0.0204 \pm 0.0006}$ & 200 & 30 \\
laser      & 993/4 &$\mathit{0.0193 \pm 0.0039}$   & 80   & 0.8913 & $\mathit{0.0204 \pm 0.0037} $& 60   & 0.5 & $ \mathbf{0.0160 \pm 0.0006} $ & 60  & 45 \\
treasury   & 1049/15 &$0.0102 \pm 0.0012          $  & 150  & 0.3162 & $0.0097 \pm 0.0009          $& 100  & 1   & $\mathbf{0.0084 \pm 0.0002}$ & 150 & 65 \\
dee        & 365/6 &$0.0486 \pm 0.0113          $  & 10   & 8.9125 & $0.0501 \pm 0.0125          $& 10   & 1   & $0.0472 \pm 0.0077         $ & 10  & 40 \\
machineCPU & 209/6 &$\mathit{0.0486 \pm 0.0115} $  & 10   & 7.9433 & $0.0398 \pm 0.0104          $& 16   & 0.5 & $\mathbf{0.0290 \pm 0.0023}$ & 26  & 60 \\
kin8nm     & 8192/8 &$\mathit{0.0654 \pm 0.0024} $  & 1000 & 0.0631 & $0.0636 \pm 0.0013          $& 1000 & 1   & $\mathbf{0.0514 \pm 0.0009}$ & 900 & 20 \\
\bottomrule
\end{tabular}
\end{center}
\end{table*} 

\begin{figure}[t]
	\centering
\includegraphics[width=0.158\textwidth]{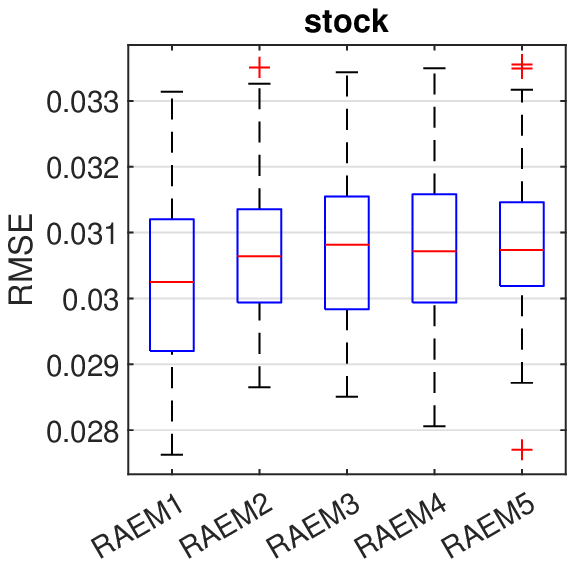}
\includegraphics[width=0.158\textwidth]{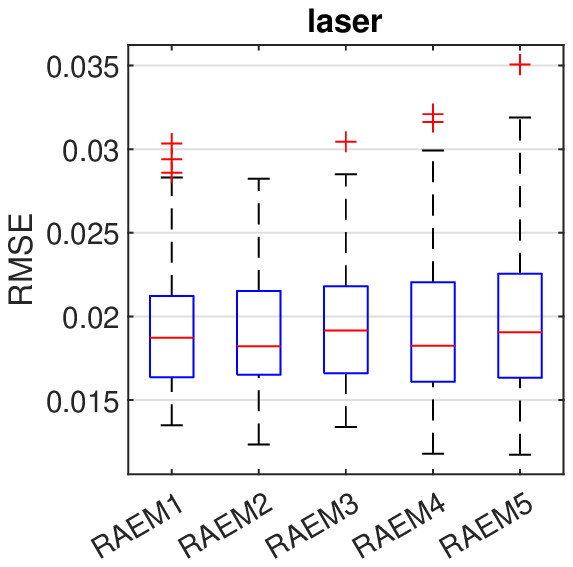}
\includegraphics[width=0.158\textwidth]{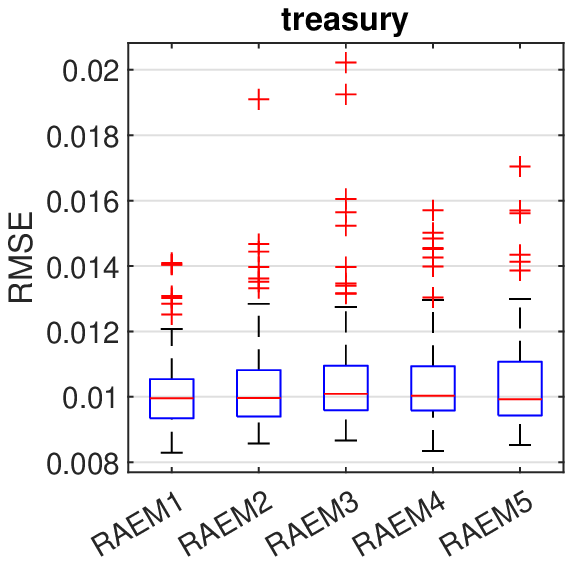}
\includegraphics[width=0.158\textwidth]{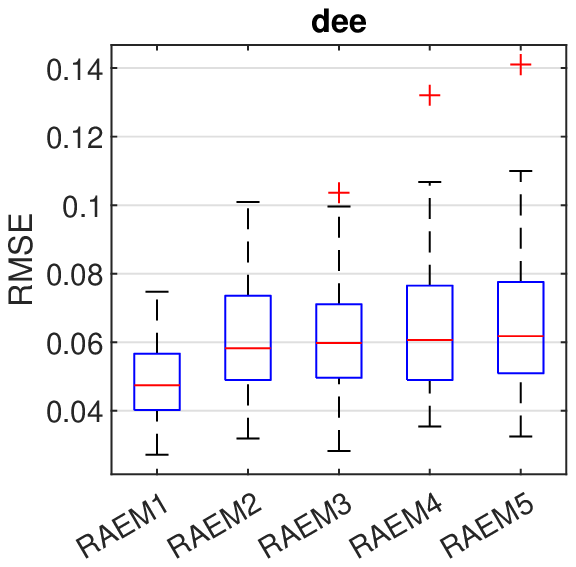}
\includegraphics[width=0.158\textwidth]{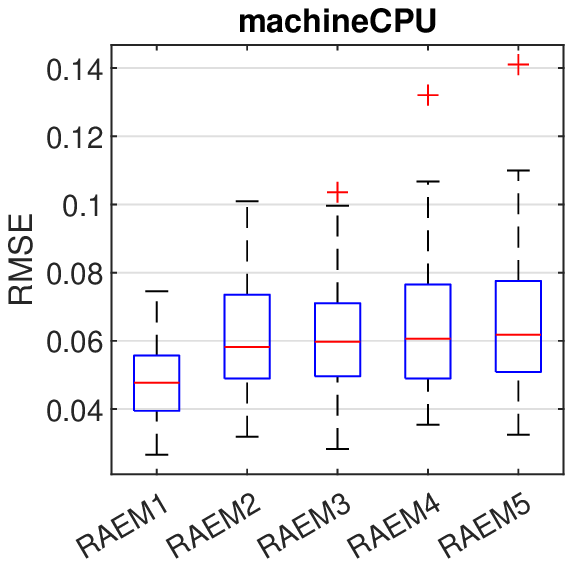}
\includegraphics[width=0.158\textwidth]{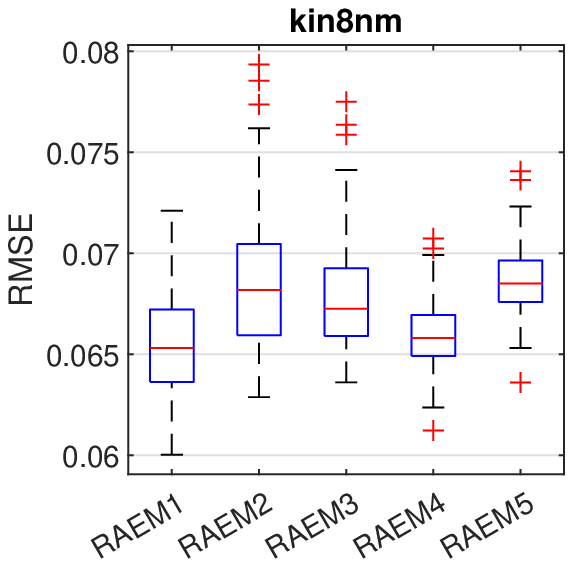}
\caption{Comparison of RAEM variants.}
	\label{figHist}
\end{figure}

Fig. \ref{figHist} compares variants of RAEM:\\ 
\textbf{RAEM1} the improved RAEM proposed in this study (optimized interval for weights $w$, $U_{AE}=[-u_{AE}, u_{AE}]$; biases $c$ and $b$ determined from \eqref{eqc} and \eqref{eqDer5a}, respectivelly),\\
\textbf{RAEM2} variant with the fixed interval for $w$, $U_{AE}=[-1, 1]$; biases $c$ and $b$ determined from \eqref{eqc} and \eqref{eqDer5a}, respectivelly,\\
\textbf{RAEM3} variant with random selection of both $w$ and $c$ from the fixed interval $[-1, 1]$; biases $b$ determined from \eqref{eqDer5a},\\
\textbf{RAEM4} variant with random selection of $w$, $c$ and $b$ from the fixed interval $[-1, 1]$,\\
\textbf{RAEM5} variant with random selection of both $w$ and $c$ from the fixed interval $[-1, 1]$; biases $b$ determined as mean values of weights, $b_i = \overline{a}_i$.

Note that the proposed modification of RAE in most cases gave the lowest errors compared to other RAE variants. The laser and treasury data sets were insensitive to the method of generating weights and biases. For these data sets, the errors for all RAE variants were similar.  

\section{Conclusion}

In this work, we showed that RAE based randomized learning of FNN suffers from several drawbacks. First, RAE produces the random weights for the FNN predictive model by taking into account just the input data. This approach is questionable because the input data does not contain information about TF complexity. TFs with strong fluctuations need higher weights than flat TFs to be modeled accurately. Unfortunately, RAE for both these cases can generate similar sets of weights ignoring completely TF complexity. Second, RAE does not generate the biases for the FNN predictive model. These biases, which determine the distribution of the activation functions in the input space, are crucial for the  approximation properties of the predictive model.

In this study, we propose improved, unsupervised parameter learning using RAEs. First, we introduce the possibility of controlling the magnitude of the random weights produced by RAE. This is realized by appropriately generating the RAE hidden node parameters. Second, we determine the biases for the FNN predictive model so that the sigmoids have their steepest fragments introduced into an input hypercube. These fragments are the most useful for modeling TF fluctuations. The proposed modifications make the RAE method more flexible, more data dependent and more dependent on the complexity of the solved problem.

The experimental part of the work does not provide evidence that the improved RAEM outperforms in fitting accuracy other new methods of generating random parameters. Moreover, its complexity is much greater as it requires additional learning of RAE. Therefore, applying it to regression problems, rather than simpler and faster methods, may be questionable.   

\end{document}